\documentclass[12pt, a4paper]{article}
\usepackage{graphicx}
\usepackage{natbib}
\usepackage{url}
\usepackage{setspace}

\title{How Do I Know What to Say Next?\\
\large Barenholtz's Autogenerative Theory as an Enrichment\\of Harrisean Integrationism}
\author{Prof J. Mark Bishop\\
\small Emeritus Goldsmiths, University of London
\and
Prof Stephen J. Cowley\\
\small Emeritus University of Southern Denmark}

\date{June 2026}

\begin{document}

\maketitle

\begin{abstract}
Roy Harris's Integrationist linguistics offers a compelling critique of the referentialist tradition embedded deep at the heart of computational approaches to language, arguing that language is not a code that maps onto a pre-given world but a situated, bipartite activity oriented toward prospective joint action. Yet Integrationism leaves certain explanatory gaps: it does not fully account for the structural mechanism by which signs sustain prospective openness, it undertheorises the continuity between linguistic and non-linguistic semiotic activity, and it offers no detailed account of the structural properties of the accumulated archive of past integrations. This paper argues that Elan Barenholtz's autogenerative theory of language, developed in response to the behaviour of Large Language Models (LLMs), can fill precisely these gaps, enriching Integrationism without undermining any of its core commitments. Specifically, the autogenerative account provides: a structural mechanism for the prospective openness that Harris identifies as central to bipartite communication; a computational correlate for Harris's thesis of semiotic continuity between language and other sign-making activity; and a theory of the archive: what the accumulated residue of past integrations looks like and how new participants draw upon it. The synthesis preserves Harris's ontological primacy of the situated integrative act while adding explanatory content that Integrationism itself does not supply. For practitioners and researchers in natural language processing and large language model design, the argument offers a principled account of what the statistical structure that LLMs so effectively exploit actually is, and of what it cannot, by its nature, provide.
\end{abstract}

\sloppy

\section{Introduction}
Roy Harris, who served as Professor of General Linguistics at Oxford until his death, developed Integrationist Linguistics over several decades as a sustained critique of what he called the ``language myth,'' namely the assumption, embedded so deeply in Western thought that it is rarely noticed, that words are fixed labels transparently conveying thoughts about an independent external world.~\citep{Harris1981} On the ``telementational'' view Harris attacked, communication succeeds when a thought in one mind is successfully replicated in another: language is merely the transparent conduit. Against this, Harris argued that language is a situated, bipartite activity: signs do not carry fixed meanings between speakers but coordinate what participants will do \emph{next}, together, in a particular context. Meaning is not transmitted; it is prospectively constructed.

In a recent \textbf{iai} essay, cognitive scientist Elan Barenholtz argues that Large Language Models (LLMs) have revealed something equally radical about the nature of language: that it does not work by pointing to or describing reality, but by generating contextually appropriate continuations of itself.~\citep{Barenholtz2026} Barenholtz calls this the ``autogenerative property'' of language. Crucially, LLMs operate without any direct exposure to the lived world (the proverbial ``deaf, dumb and blind kid, who sure plays a mean pinball,'' to borrow a phrase from The Who~\citep{TheWho1969}) yet nevertheless generate coherent and contextually appropriate language. An LLM knows nothing about the actual redness of \emph{red} or the physical distance of \emph{far}; it knows only where these words fall in relation to every other word in a high-dimensional relational space. And yet from this manipulation of what he calls ``empty symbols,'' coherent, contextually appropriate language emerges.

Barenholtz is careful, however, not to claim that the human brain literally \emph{is} an LLM. The argument is more subtle: the success of LLMs reveals a structural property present in the accumulated corpus of human linguistic activity, and the brain (as a situated, embodied, living organ) may exploit that property without implementing it computationally. As Barenholtz puts it, ``it's not that the brain is literally an LLM. Rather, the proposal is that it too may exploit the structure we now know is already present in language, generating words based on the predictive structure of previous words.'' This distinction is critical for the argument that follows. The enrichment of Integrationism we propose is not a reduction of language to computation; it is a claim about the statistical structure of the corpus (the accumulated residue of past linguistic acts) that both LLMs and human speakers, in their very different ways, are able to exploit.\footnote{A terminological caution is necessary here, and bears on all that follows. When we use the phrase ``structure of language'' or ``property of language,'' we do not mean to reify language as an abstract, pre-given entity, a mistake Harris spent his career opposing. Following the reading of Barenholtz developed in \S5, such phrases should be understood as shorthand for \emph{the statistical structure of the corpus}: the accumulated, historically contingent precipitate of past integrative acts. There is no ``language itself'' floating above these acts; there is only the structural trace they leave behind. This point is further developed from the radical linguist's side in \citet{Cowley2026}, where it is argued that LLMs use data, not `language'. }

The two accounts have been read primarily as convergent critiques of referentialism, and the convergence is real and substantial. But the more productive question is whether Barenholtz's account can do something that the convergence literature has not yet fully exploited: not merely \emph{agreeing} with Harris, but \emph{enriching} him, supplying mechanism, continuity, and archive theory where Integrationism, for all its philosophical power, leaves explanatory gaps. This question has direct implications for AI and computational linguistics: if the enrichment succeeds, it provides a theoretically grounded account of what large language models actually model, what they cannot capture, and why the gap between statistical competence and situated linguistic activity is not merely a matter of scale or data.

This paper argues that it can, and that the enrichment operates along three distinct axes. First, the autogenerative account provides a structural mechanism for the prospective openness that Harris identifies as central to the bipartite sign but never fully explains (\S3). Second, Barenholtz's multimodal extension of the autogenerative framework, encompassing imagery and perception alongside language, provides computational support for Harris's own thesis of semiotic continuity between linguistic and non-linguistic activity (\S4). Third, the autogenerative account can be read as a theory of the archive: a characterisation of what the accumulated residue of past integrations structurally is, and how new participants draw upon it, a topic Harris acknowledged but left undertheorised (\S5). Throughout, care is taken to ensure that the enrichment operates within, rather than against, the core commitments of Integrationism.

It is worth noting at the outset that the argument developed here has been reached, independently, by researchers working from substantially different starting points. Barenholtz~\citep{Barenholtz2026} approaches the question as a cognitive scientist analysing LLM behaviour; \citet{Cowley2026} approaches it as a radical ecolinguist working in the tradition of distributed language; and Pinna~\citep{Pinna2026} approaches it from a philosophy of science perspective, arguing that LLM competence arises from reconstructing norm-shaped continuations rather than from manipulating representations directly tied to reality, and characterising the corpus as preserving ``sedimented traces'' of norm-governed practices: an intuition strikingly convergent with the archive account developed in \S5. The fact that four independent lines of inquiry converge on the autogenerative capacity as central to understanding the relationship between LLMs and language is itself significant evidence for the account's validity, and situates the present paper within a broader research programme rather than as an isolated proposal.\footnote{The fourth line of inquiry is the present paper itself~\citep{BishopCowley2026}. The engagement with Harris's Integrationism, which forms its philosophical core, grew out of Bishop's prior work on the relationship between AI and the philosophy of language. Cowley initially entered the project as a commentator on an earlier draft, and his contribution proved sufficiently generative that he subsequently became a co-author. The independence of the four accounts is therefore genuine, albeit that two of the four have since converged in the writing of this paper.}

\section{The Integrationist Foundation}
Harris's central claim, developed across \emph{The Language Myth}~\citep{Harris1981} and \emph{Reading Saussure}~\citep{Harris1987}, is that the Western linguistic tradition has systematically misconstrued language by treating it as a fixed code shared between speakers.\footnote{It is worth noting that the ``language myth'' as Harris construes it has two distinct components: the \emph{code model} (the view that linguistic forms map onto fixed meanings) and the \emph{telementational model} (the view that communication succeeds when a thought in one mind is replicated in another). Our paper focuses primarily on the telementational aspect, since this is most directly relevant to the LLM case; but the code model is an equally important target of Harris's critique, and the two are not always distinguished in the literature.} The Saussurean model, whatever its merits as a structural account, inherits and codifies this misconception: it posits a \emph{langue}, an abstract system floating above individual acts of communication, and treats the individual utterance as a mere instantiation of that system. For Harris, this gets things exactly backwards. There is no langue; there is only the individual, creative, situated act of integration through which participants make signs mean something, for themselves, here, now.

Several commitments follow from this. The sign is \emph{indeterminate}: no sign carries a fixed, context-independent value. Meaning is not stored inside words and retrieved; it is constructed afresh in each communicative situation by participants who bring their own histories, bodies, and purposes to the encounter. Language is therefore characteristically \emph{bipartite}: in its paradigm cases it involves at least two parties jointly orienting toward a shared situation, and its constitutive feature is the coordination it enables between them. Harris does, however, allow for self-communication, in which a single individual integrates signs without a second participant; bipartite coordination is thus the paradigm of signmaking rather than its necessary condition. Communication is \emph{prospective}: the function of the sign is not to report a state of affairs but to coordinate what the participants will do \emph{next}, together. And language is \emph{embodied}: the sign-maker is not a disembodied processor but a biomechanical agent whose bodily situation is constitutive of the sign itself, not merely its vehicle.

These commitments are not independent theses loosely bundled together; they form a coherent architecture. Indeterminacy explains why bipartite coordination is the paradigm case of signmaking: because signs have no fixed values, each integration constructs its meaning afresh, uniquely and non-transferably, in the encounter between the sign-maker and the situation. This is not to say that meanings are shared or converged upon between participants; Harris's indeterminacy thesis rules out any such convergence. Each integration remains irreducibly singular, even when it produces behaviour that appears coordinated. Bipartite activity explains why communication is prospective: what the participants are jointly doing is not exchanging information but coordinating action. And embodiment anchors the whole account in the specificity of situated human practice.

What Integrationism does not provide, however, is a detailed account of the \emph{mechanism} by which signs sustain prospective openness: why it is that signs structurally \emph{invite} further activity rather than closing it down. Nor does it offer a systematic account of semiotic continuity beyond the general claim that linguistic and non-linguistic sign-making are of a piece. And it has relatively little to say about what the accumulated archive of past integrations is, structurally, and how participants draw on it. These are not fatal gaps; Integrationism is a philosophical account of what language is, not a cognitive science of how it works. But they are genuine lacunae, and it is here that Barenholtz's autogenerative account has something to offer. For the AI and computational linguistics reader, these lacunae map onto familiar open questions: why do language models that lack any grounding in the world nevertheless produce contextually appropriate outputs; what is the relationship between the vector space representations that LLMs learn and the meaning that human speakers construct; and what does the success of LLMs tell us about the structure of the linguistic resources human speakers draw on.

One further clarification is necessary before proceeding. The enrichment proposed here goes, at certain points, beyond what Harris himself explicitly argued. Harris's project was primarily critical and philosophical: he dismantled a misconception but left the positive account of what language can become, given how the archive of past integrative acts has been structured, underspecified. The sections that follow develop that positive account using Barenholtz's autogenerative framework. This is a deliberate extension of Harris rather than a misreading of him, and it is offered in that spirit.

\section{Autogeneration as the Mechanism of Prospective Openness}
Harris's bipartite model holds that the sign coordinates future joint action between participants. The sign does not move content from one head to another; it opens a horizon of possible next moves that participants jointly navigate. This is what Harris means by the prospective character of communication: the sign is not a report but an orientation, not a description but an invitation to act.

But why, structurally, should the sign have this character? Harris's answer is philosophical: because signs have no fixed referential content to transmit, they cannot function as packets of information. Their function must therefore be something else, and that something else is the coordination of prospective activity. This is compelling as far as it goes, but it leaves the mechanism of prospective openness unaccounted for. Why does the sign open a horizon rather than simply failing to close one?

Barenholtz's autogenerative account offers a structural answer. An LLM trained on the accumulated residue of human linguistic activity learns, in effect, that every element of the corpus exists in a web of probabilistic relations with every other element. No word has a fixed ``address content''; each is defined entirely by its relational position in a high-dimensional space, weighted by the statistical history of its co-occurrences. What this means is that every element of the corpus is structurally \emph{open}: it does not point to a fixed referent but generates a probability distribution over possible continuations.~\citep{Barenholtz2026} This is not a limitation of the corpus but its constitutive feature: what Barenholtz calls the autogenerative property. As argued in \citet{Cowley2026}, the LLM does not thereby ``talk'' in any sense that entails languaging: it generates what is there termed \emph{aggregated patternings}, statistically normalised outputs that trigger epicommunicative effects in human readers, rather than producing linguistically situated, embodied, prospective acts of communication.

Translated into Harrisean terms: the sign is prospectively open \emph{because} it is autogenerative. Its relational, non-referential structure means that it does not terminate in a fact about the world but generates a range of possible next moves (linguistic, imagistic, and behavioural) that participants must jointly resolve in their particular situation. The chair example that both thinkers independently reach is instructive. The utterance ``there is a chair in the living room'' does not mean anything in isolation; it generates a distribution of possible responses (moving the chair, noting its presence, asking why it matters) that only the specific bipartite context can narrow down. Barenholtz's formulation, ``conditions beget conditions,'' is a precise structural description of what Harris observes as prospective coordination.

This enriches rather than challenges Integrationism because it adds mechanism to a philosophical observation without disturbing the observation itself. Harris is right that the sign coordinates prospective joint action; Barenholtz explains, at the level of linguistic structure, why the sign is built for this rather than for anything else. The enrichment is explicitly non-reductive: the autogenerative mechanism does not replace the bipartite integrative act but explains why that act must be bipartite in the first place. A sign with a fixed referential address could, in principle, be decoded unilaterally; an autogenerative sign, with no fixed address and a distribution of possible continuations, \emph{requires} a second participant to resolve its indeterminacy in a particular situation. The mechanism and the philosophical account are not in tension; they support each other.

One caveat is necessary. Harris would resist any framing that locates the autogenerative structure ``in language'' as an abstract system independent of integrative activity. The move must be considered with care: the autogenerative property is not a feature of \emph{langue} but a structural tendency that \emph{emerges from} the accumulated history of integrative acts, visible in retrospect in the corpus. It is not prior to use but constituted by use, which is, of course, exactly what Harris himself would want to say about any genuine property of language. For the LLM researcher, this reframing is significant: the high-dimensional vector space that a language model learns is not a representation of language as an abstract system but a statistical precipitate of the history of human communicative acts. This is why such models can be extraordinarily effective at generating contextually appropriate continuations while remaining, as \citet{Cowley2026} puts it, wholly reliant on structured information rather than on the situated, prospective, bipartite activity that constitutes language proper.

It follows that the autogenerative process, understood in this way, is strictly a functional mechanism: it characterises a structural property of the corpus independently of any particular implementation, any particular person, and any particular environment. It should be noted that recursivity as such is not unique to language; it is a feature of all coordinated activity. What the autogenerative account specifically reveals is the presence of this recursive, open-ended structure in the statistical organisation of the linguistic corpus in particular, and it is this that LLMs bring into view. As such, and as all four independent accounts noted above agree, it is not sufficient to account for any act of languaging. As Cowley observes (personal communication), it cannot account for even the simplest communicative act, such as saying ``boo to a goose''. The autogenerative mechanism explains the openness of the sign; it does not explain what participants do with that openness in the living moment of integration.

\section{Autogeneration and Semiotic Continuity}
One of Harris's most important and underappreciated arguments, and one with direct implications for the design and interpretation of multimodal AI systems, concerns the continuity between linguistic and non-linguistic semiotic activity. Against the tradition that treats language as a sui generis faculty, radically different in kind from other forms of human sign-making, Harris argues in \emph{Signs, Language and Communication} that there is no principled boundary between linguistic and non-linguistic signs.~\citep{Harris1996} Both are forms of integration; both are situated, embodied, and prospective in their orientation. The sharp boundary drawn by the Chomskyan tradition between a species- specific language faculty and general cognition is, for Harris, merely another artefact of the language myth, a consequence of treating language as a fixed code rather than a form of activity. This boundary is most explicitly drawn in the influential paper by \citet{HauserChomskyFitch2002}, which distinguishes a ``Faculty of language---narrow sense (FLN)'', hypothesised to be uniquely human, from broader sensory-motor and conceptual-intentional systems shared with other species, precisely the compartmentalisation that Harris's continuity thesis rejects.

Barenholtz's account supports this thesis in a way that Harris could not have anticipated, because it draws on empirical evidence that did not exist when Harris was writing. Barenholtz notes that the autogenerative property of language is not confined to text: multimodal AI systems, trained jointly on language and images, exhibit the same structural openness across modalities. An image-generating model does not map a phrase onto a fixed visual referent; it generates a distribution of possible images that are contextually appropriate continuations of the linguistic input. Conversely, an image-captioning model does not read off a fixed verbal description of an image; it generates a distribution of possible linguistic continuations of the visual input.~\citep{Barenholtz2026} The autogenerative structure, in other words, is not peculiar to language: it is a feature of the entire semiotic system within which language participates.

This is precisely what Harris's semiotic continuity thesis predicts. If linguistic and non-linguistic signs are both forms of integration, and if integration is constitutively prospective and open, then we should expect the structural openness that Barenholtz identifies in language to be present across the semiotic field. The LLM evidence does not establish this continuity (Harris's philosophical argument already does that), but it provides a striking computational correlate. The same statistical architecture that produces open, contextually appropriate linguistic continuations also produces open, contextually appropriate imagistic ones. The boundary between modalities, like the boundary between linguistic and non-linguistic signs, turns out to be less sharp than the traditional picture assumes.

There is a further consequence worth drawing out. Barenholtz's multimodal account includes perception and motor behaviour within the autogenerative loop: language generates conditions, conditions generate imagery, imagery is compared with perceptual input, and motor behaviour follows. This maps naturally onto Harris's insistence that embodiment is constitutive of the sign rather than merely its vehicle. The body in Barenholtz's account is not a passive executor of linguistically generated instructions; it is a participant in an ongoing generative process that includes perception, action, and the feedback between them. This is not quite the biomechanical primacy that Harris insists on: Harris would want to say that the body is \emph{prior} to language, not merely included within the generative loop, but it moves in the right direction and is considerably more compatible with Integrationism than a purely disembodied computational account would be.

A note of caution is warranted, however, about the strength of the LLM evidence for Harris's continuity thesis. From the radical linguist's perspective developed in \citet{Cowley2026}, one would resist the inference that multimodal LLM outputs \emph{demonstrate} the continuity between linguistic and non-linguistic human sign-making: such outputs are still aggregated patternings triggering epicommunicative effects in human readers, and their cross-modal character tells us about the structure of the corpus, not about embodied semiotic activity. Our claim here is therefore the more cautious one: the multimodal behaviour of LLMs is a \emph{computational correlate} of Harris's continuity thesis, a structural parallel that is suggestive rather than probative. Harris's own philosophical argument for continuity does not depend on it and stands independently.

What the LLM evidence does illuminate is the domain of what we call \emph{slow languaging}: the supra-individual, historically accumulated patterns of verbal and non-verbal practice that have discoverable statistical properties, precisely those that LLMs reveal. Slow languaging is where Harris's strict situatedness can be partially relaxed: certain languaging practices \emph{are} supraindividual and have structural features that both constrain and afford the living integrative act. The multimodal autogenerative account describes the archive of slow languaging across modalities, which is a genuine and useful contribution, even if it falls short of establishing Harris's full continuity thesis.

One important limitation of the present account must be acknowledged here. Both Barenholtz's framework and our reading of it tend to privilege the distributed resources of the population (the corpus, the archive, the statistical traces of past languaging) over the active contribution of the individual brain. \citet{Cowley2026} argues that human brains use something like a high-dimensional vector space, but crucially go beyond it: through judging, attending, predicative activity, and a life history of coordinated engagement, human speakers actively add value to the structured information the archive affords in ways that LLMs cannot. Harris himself tends to play down the biomechanical brain in favour of the sign-maker understood holistically, but a fully adequate account of the relationship between archive and living act will require a richer treatment of these individual, brain-based powers of judgement and predicative activity than either Harris or Barenholtz supplies. Crucially, as \citet{Cowley2026} observes, what distinguishes human languaging from LLM output is not merely that humans add value to statistical patterns but that they can be specifically and appropriately \emph{right}: not simply contextually fitting, but exercising judgement in ways that are accountable to the world and to other participants in a way no archive-driven process can be. This is not a deficiency the present paper can remedy: it is a direction that a more complete synthesis must pursue, one that takes the enlanguaged brain seriously as an active participant in the integrative encounter rather than merely a conduit between the archive and the body.

\section{Autogeneration as a Theory of the Archive}
Harris was well aware that language leaves traces. Texts, inscriptions, recordings, and other artefacts persist beyond the integrative acts that produced them and can be taken up by new participants in new situations. For the computational linguist, this observation bears directly on the nature of the training corpus: the data on which a large language model is trained is precisely such a trace, a vast archive of the residue of past human communicative acts, structurally rich but ontologically distinct from the living activity that produced it. He was careful to insist that these traces are not language itself; language proper arises in the situated integrative act, not in its residue; he offered no detailed account of what the archive is, structurally, or of how participants draw upon it when they integrate.

This is a genuine gap. Integrationism's insistence on the primacy of the situated act is philosophically important and must be preserved, but it leaves open the question of what participants are actually doing when they exploit the accumulated resources of their linguistic community. When a speaker reaches for a word, they are drawing on something that precedes the current act of integration (not a fixed code, but not nothing either). Harris's account gives us no detailed picture of what that something is.

Barenholtz's autogenerative account can be read as precisely this: a theory of the archive. What the accumulated residue of past integrations looks like, structurally, is a vast relational space in which every element is defined by its statistical relations to every other, a space with no fixed address contents, no stored meanings, but a rich and highly structured set of probabilistic tendencies. New participants do not retrieve fixed meanings from this space; they exploit its tendencies, drawing on the statistical history of past integrative acts to generate contextually appropriate continuations in their current situation.

Crucially, this framing preserves exactly the distinction Harris needs. The archive is not language; it is the structural precipitate of past languaging, the trace that integrative activity leaves behind. Language proper remains what Harris says it is: a living, situated, bipartite act of integration. But the archive is not, therefore, structurally inert. It has determinate statistical properties, precisely those that Barenholtz's autogenerative account characterises, and new integrators exploit these properties in their acts of sign-making. The archive is, so to speak, the resource that participants bring to the integrative encounter, without being the encounter itself.

The precise nature of the archive's influence on the living act requires care. \citet{Cowley2026} adopts a position between Harris and Barenholtz on this question, concurring that the archive is a structure of statistical relations, an aggregate of aggregated patternings, that does not bear on acts of contextualisation except as \emph{affording constraints}. But, unlike Harris, that account accepts that human doings can be partly described by rule-following, and that we can both draw on and discover supracommunicative domains of normativity that \emph{influence} our languaging without determining it. In those terms, the archive has a constitutive role in ``language games'' and most strongly in mathematical ones, a point that connects to the notion of `slow languaging' indicating how verbal practices acquire supra-individual, discoverable structure. This middle position is productive for our argument: it allows us to say that Barenholtz's autogenerative account characterises the archive as a set of affordances, statistical tendencies that constrain and enable the living integrative act without replacing or reducing it.

Two further refinements to the archive metaphor are necessary here. First, the archive does not consist of ``past language'': it consists of the traces of past languaging, the aggregated patternings and indices of tokenizings that prior integrative acts have left behind. This distinction, emphasised in \citet{Cowley2026}, is not merely terminological. To speak of ``past language'' is to risk reifying the very entity whose existence Harris denies; to speak of traces of past languaging is to remain faithful to the ontology of the integrative act while acknowledging that those acts leave a structurally significant residue. Pinna's formulation, ``sedimented traces'' of norm-governed practices~\citep{Pinna2026}, captures the same intuition from a philosophy-of-science perspective and aligns with our own account of the corpus as an accumulated residue. Second, the relationship between the archive and the living act is not one-directional. Participants do not merely draw on the archive; they continuously remake it through their acts of languaging. As Cowley likes to tell his students: languaging is like the sea in which we swim, but we are unusual fish because we make up the sea. The archive is not a fixed resource standing over against the speaking subject; it is a precipitate of human activity that is itself continuously reshaped by that activity. This dialectical relationship is important for the present account: it means that the statistical structure Barenholtz's autogenerative account reveals is not a permanent or context-independent feature of some abstract system, but a historically contingent and continuously evolving precipitate of the languaging practices of a community.

This reading is further supported by Barenholtz's own opening anecdote, in which archaeologists unearth clay tablets from a lost civilisation and discover that the symbols are self-predicting, such that the sequence on one part of the tablet is mathematically sufficient to derive what appears on another:

\begin{quote}
``The finding that the symbols contain this predictive structure would be an
extraordinary insight. But we still might ask, `what do the symbols mean?'\thinspace ''
\end{quote}

Barenholtz raises this question in order to press the autogenerative account further. But from a Harrisean perspective, the anecdote maps perfectly onto the archive distinction: the tablets are the archive, and their autogenerative structure is a genuine and important property of that archive. What it cannot tell us is what the symbols \emph{meant}, since meaning pertains to the situated integrative acts of the civilisation that produced them, acts to which the archaeologists have no access. The autogenerative structure of the archive and the living meaning of the integrative act are genuinely different things, and Barenholtz's own anecdote, read through Harris, makes exactly this point.

\section{Conclusion}
The relationship between Barenholtz's autogenerative account and Harris's Integrationism is not merely one of convergence, though the convergence is real and substantial. Both thinkers reject referentialism; both treat language as a form of doing rather than describing; both insist that meaning is not stored in signs and that context is constitutive rather than decorative. These affinities are genuine and have been noted.

But the more productive relationship is one of enrichment. Integrationism provides an apposite philosophical ontology: language is a situated, bipartite, prospective, embodied activity, and any account that loses sight of these features has failed to capture what language is. Barenholtz's autogenerative account does not challenge this ontology; it adds to it. It provides a structural mechanism for the prospective openness that Harris rightly identifies as central to the sign but does not fully explain. It provides a computational correlate for Harris's semiotic continuity thesis, suggesting that the structural openness of language extends across the entire semiotic field of language, imagery, and perception. And it enables a theory of the archive, a characterisation of what the accumulated residue of past integrations structurally is and how new participants exploit it, that Integrationism itself leaves undertheorised.

The synthesis that emerges is philosophically modest in the best sense: it does not claim to resolve the tension between a statistical account of linguistic structure and a philosophical account of living meaning. As Barenholtz's own archaeologist anecdote implies, and as John Searle's observation that the computational simulation of a rainstorm does not make anyone wet reminds us, there remains an irreducible gap between the structure of the archive and the meaning of the integrative act.~\citep{Searle1980} What the synthesis does claim is that the two accounts are complementary rather than competing: Harris tells us what language \emph{is}; Barenholtz tells us what the archive \emph{looks like} and, in doing so, helps explain why the living act of integration must have the prospective, open, bipartite character that Harris assigns to it.

The semiotic continuity argument (\S4) points toward a further horizon that the present paper can only gesture at. \citet{Cowley2026} argues, in a Sellarsian vein, that the autogenerative process extends beyond language to what can be co-perceived: the shared, enlanguaged world that both lay people and scientists inhabit. If the ``manifest image'' (what ordinary perception makes available) and the ``scientific image'' (what science makes manifest) are both shaped by the accumulated archive of slow languaging, then the claim of semiotic continuity reaches not merely across sign-modalities but into the structure of perceptual experience itself. This is a significant extension of Harris's own continuity thesis, and one that connects the autogenerative account to long-standing questions about the relationship between language, perception, and the constitution of a shared world. It is, we suggest, the natural subject of the next paper in this research programme, one that would bring together the archive account developed here, the brain-based powers of judgement and predicative activity emphasised in \citet{Cowley2026}, and the social normativity of norm-governed continuations developed in \citet{Pinna2026}.

The age of LLMs has not, as some have supposed, rendered Harris's project obsolete. It has, if anything, made it more necessary, by providing a powerful demonstration of just how much can be achieved by exploiting the statistical structure of the archive, and therefore of just how much remains that the archive alone cannot provide. The notion of slow languaging is working to this remainder: the archive captures the supra-individual, historically accumulated structure of slow languaging practices, and LLMs reveal this structure with unprecedented clarity. But slow languaging is itself embedded within the faster, living, embodied integrative activity that Harris describes, and it is that activity, with its prospective openness, its bipartite character, and its irreducible situatedness, that the archive alone can never reconstruct. For researchers working on large language models and computational approaches to language, this is not a counsel of despair but a clarification of the problem space: LLMs are powerful precisely because the archive is rich and its statistical structure is deep. Understanding what that structure is, and what it is not, is a prerequisite for understanding both the remarkable capabilities of current systems and the principled limits that no amount of scaling can overcome.

\end{document}